\documentclass[letterpaper, 10 pt, conference]{ieeeconf}  

\IEEEoverridecommandlockouts                              

\usepackage{xcolor}
\usepackage{amsmath}
\usepackage{amssymb}
\usepackage{dsfont}
\usepackage{graphicx}
\usepackage{algorithm}
\usepackage{algpseudocode}
\usepackage{romannum}
\usepackage{hyperref}
\usepackage[export]{adjustbox}
\usepackage{subcaption}
\usepackage{pdfpages}
\usepackage{soul}

\DeclareFontFamily{OT1}{pzc}{}
\DeclareFontShape{OT1}{pzc}{m}{it}{<-> s * [1.10] pzcmi7t}{}
\DeclareMathAlphabet{\mathpzc}{OT1}{pzc}{m}{it}
\DeclareMathOperator*{\argmax}{arg\,max}

\newcommand{\revised}[1]{{#1}}

\graphicspath{{./figs/}}

\title{\LARGE \bf
Interactive Robotic Grasping with Attribute-Guided Disambiguation
}

\author{Yang Yang, Xibai Lou, and Changhyun Choi
\thanks{*This work was supported by UMII MnDRIVE Ph.D. Graduate Assistantship and MnDRIVE Initiative on Robotics, Sensors, and Advanced Manufacturing.}
\thanks{$^\dagger$The authors are with University of Minnesota, Minneapolis, USA {\tt\small \{yang5276, lou00015, cchoi\}@umn.edu}}%
}

\begin{document}

\maketitle

\begin{abstract}
Interactive robotic grasping using natural language is one of the most fundamental tasks in human-robot interaction. However, language can be a source of ambiguity, particularly when there are ambiguous visual or linguistic contents. This paper investigates the use of object attributes in disambiguation and develops an interactive grasping system capable of effectively resolving ambiguities via dialogues. Our approach first predicts target scores and attribute scores through vision-and-language grounding. To handle ambiguous objects and commands, we propose an attribute-guided formulation of the partially observable Markov decision process (Attr-POMDP) for disambiguation. The Attr-POMDP utilizes target and attribute scores as the observation model to calculate the expected return of an attribute-based (e.g., ``what is the color of the target, red or green?'') or a pointing-based (e.g., ``do you mean this one?'') question. Our disambiguation module runs in real time on a real robot, and the interactive grasping system achieves a 91.43\% selection accuracy in the real-robot experiments, outperforming several baselines by large margins. Supplementary material is available at \href{https://sites.google.com/umn.edu/attr-disam}{https://sites.google.com/umn.edu/attr-disam}.
\end{abstract}


\section{INTRODUCTION}
\begin{figure}[t]
  \centering
  \includegraphics[width=0.47\textwidth]{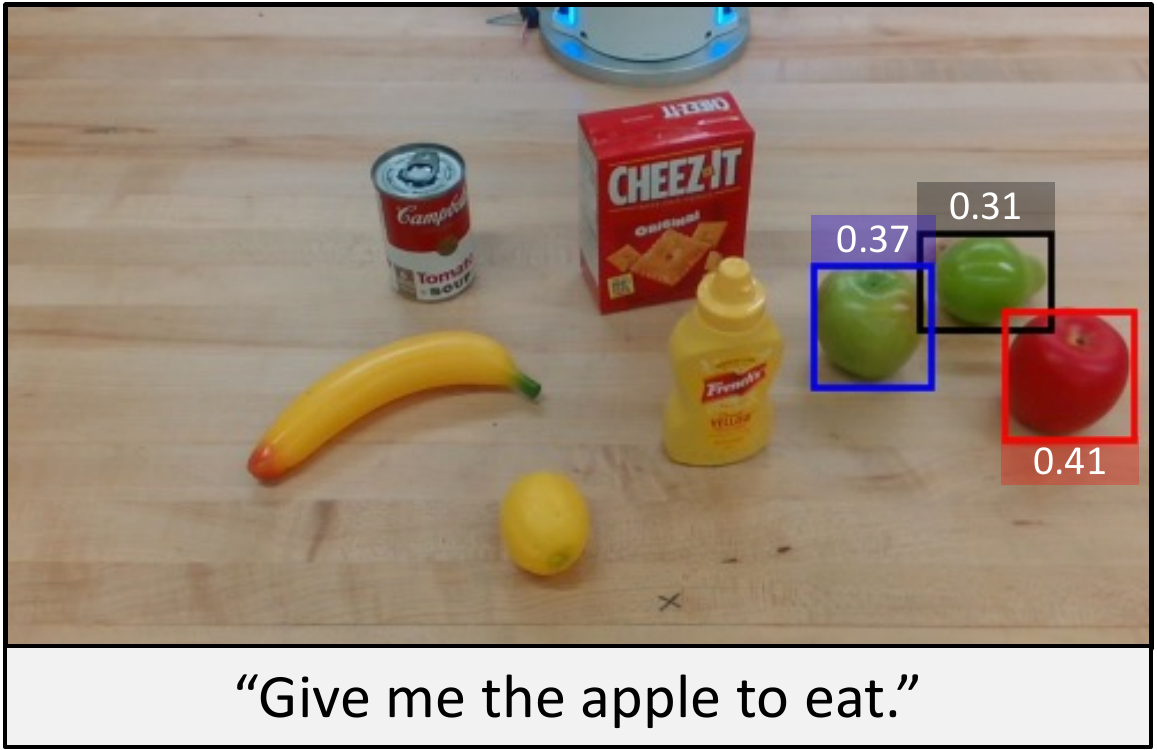}
  \caption{\textbf{Example of ambiguity in robotic grasping} with the target matching scores visualized. \revised{The robot is asked to retrieve a target object following the language command, but an ambiguity arises (i.e., the three fruits with similar scores).}}
  \label{fig:intro}
  \vspace{-10pt}
\end{figure}

As robots become more common in human-centered environments, there is a greater need for natural and effective human-robot interaction. Natural language is attracting more attention because it provides a powerful interface \revised{in robotic manipulation~\cite{tellex2020robots}, such as instructing a robot to grasp a target object.} Language, on the other hand, is a source of ambiguity. As shown in Fig. \ref{fig:intro}, an agent in an open world inevitably comes across some ambiguous visual (e.g., similar or unknown objects) or linguistic (e.g., unclear commands) contents~\cite{yang2018visual}. It is natural to consider whether the agent can simply request more information from the human user to resolve ambiguities~\cite{tellex2014ask}. As attributes (e.g., color, location) are middle-level abstractions of object properties~\cite{yang2021attribute}, they can provide critical guidance in disambiguation. This hypothesis motivates interactive robotic grasping with an attribute-guided disambiguation capability.

Recent advances in vision-and-language tasks, such as language grounding~\cite{steels2012language} and image captioning~\cite{kiros2014multimodal}, have been applied to robotic manipulation and human-robot interaction~\cite{hatori2018interactively, ahn2018interactive, mees2021composing}. In the case of an ambiguity in user commands, typical captioning-based disambiguation approaches formulate a question by greedily selecting a crop caption, lacking sequential decision-making for grasping and asking actions. Moreover, the language generation models, pre-trained on known objects, exhibit limited generalization in the real-robot tasks and even generate complex and inaccurate questions for users, causing further confusion~\cite{hossain2019comprehensive}. In contrast, our approach formulates simple yet discriminative questions regarding generic object attributes and object confidence. The proposed attribute-guided POMDP (Attr-POMDP) sequentially plans the asking and grasping actions, calculating the expected return for each action and accounting for object uncertainties.

Disambiguation in robotic manipulation~\cite{lutkebohle2009curious} is a process involving object grounding, ambiguity estimation, question generation, and action planning. It is challenging due to the difficulties in 1) object grounding in real-robot settings, 2) generating discriminative questions (i.e., what to ask), 3) handling the trade-off between asking and grasping (i.e., deciding when to ask or grasp), and 4) generalization to novel settings and unknown objects.

In this paper, we propose an interactive robotic grasping system with attribute-guided disambiguation. The key aspects of our system are
\begin{itemize}
    \item A robust objectness detector, which generalizes to unknown objects by leveraging RGB-D images, is used to annotate the objects of interest.
    \item We train an object grounding module to predict the target score (corresponding to a language command) and generic object attributes of each candidate object. The grounding results summarize scene uncertainties.
    \item For efficient disambiguation, we design an action space of attribute-based and pointing-based questions that are robust in real-robot tasks.
    \item The proposed Attr-POMDP model sequentially plans the asking and grasping actions by using the grounding results as the observation model. The goal of the planner is to balance between obtaining more information from users and the cost of asking.
\end{itemize}
Our interactive robotic grasping system can understand various command languages and will ask questions if there are any ambiguities. The attribute-guided disambiguation module can handle unclear commands and similar or unknown objects, significantly improving the grasping accuracy. Fig. 1 shows an example of ambiguity in robotic grasping, wherein our approach can successfully distinguish the target object by attribute (e.g., color) and pointing questions.

\textbf{Contributions}: This paper presents two core technical contributions: 1) a POMDP planner guided by generic object attributes that efficiently disambiguates visual and linguistic contents that are ambiguous; 2) an interactive robotic system that robustly localizes and grasps the target object via human-robot interaction. With visual observations from an RGB-D camera, our real-robot system can interactively grasp a target object following a user's command.

\section{RELATED WORK} \label{sec:related}
As robots become more integrated into people's daily lives~\cite{kroemer2021review}, we expect more powerful interfaces than keyboards, mice, and touch screens~\cite{goodrich2008human}. Natural language, as opposed to traditional tools, enables more flexible and intuitive human-robot interaction~\cite{tellex2020robots}. Language understanding tasks~\cite{forbes2015robot}, in which a robot must understand users' referring expressions~\cite{qiao2020referring} to act in response, have been proposed and researched. However, ambiguities arise when a robot fails to ground referring languages on visual concepts~\cite{li2016spatial} as a result of unclear commands, similar objects, novel targets, and so on. Our interactive grasping system is capable of understanding different types of information (subject appearances, object locations, and spatial relationships) in languages and asking discriminative questions when ambiguity is present.

To ask a meaningful question for disambiguation, we might consider greedy-based approaches such as selecting the most uncertain word~\cite{tellex2013toward, thomason2019improving}, the most contrasting attribute concept~\cite{li2017learning}, or the most likely object~\cite{mees2021composing} to ask about. Tellex \emph{et al.}~\cite{tellex2013toward} proposes an uncertainty measure for words in a grounding graph and asks a question about the most uncertain word (e.g., ``what does the word box refer to?''). To find an attribute concept that best differentiates the objects, we can calculate the attribute score contrast~\cite{li2017learning, morohashi2019query} or train a classifier~\cite{ahn2018interactive}. Mees \emph{et al.}~\cite{mees2021composing} propose to generate the least ambiguous expression about the most likely object by performing beam search on a pre-trained language generator. While these greedy-based methods can elicit more information by a single question, they lose global optimality when multiple questions are required for complex scenarios. Furthermore, they must rely on a target score threshold to decide whether to grasp or keep inquiring.

Recently, disambiguation approaches by POMDP planning have been proposed. POMDPs can help with decision-making in noisy and partially observable environments. When it comes to disambiguation, a key aspect of POMDP planning is balancing between the benefits of obtaining additional information (via asking questions) and the cost of doing so. Whitney \emph{et al.}~\cite{whitney2017reducing} forms FETCH-POMDP for disambiguation, but it relies solely on fixed questions (e.g., ``do you mean this one?'' together with a pointing gesture). In~\cite{shridhar2020ingress, zhang2021invigorate}, INGRESS-POMDP is applied for disambiguation in addition to their matching modules. INGRESS-POMDP extends FETCH-POMDP by replacing the fixed questions with the referring expressions (generated by pre-trained language generators) for each candidate object. However, the difficulty of modeling user responses is a common issue for these two POMDP models. When users are asked unguided questions, either fixed or generated, they are more likely to rephrase their desire using any words, greatly expanding the sampling space of the POMDP models. As a result, these two POMDP approaches have to exclude other candidate objects by either only modeling yes/no responses (any correcting response is discarded)~\cite{shridhar2020ingress} or sampling a single word from the whole dictionary~\cite{whitney2017reducing}. In contrast, we propose using target and attribute scores as the observation model to calculate the expected return of an attribute-based or a pointing-based question, which makes our POMDP model substantially more accurate and efficient.

\section{PROBLEM FORMULATION}
The language-instructed robotic grasping problem in this paper is stated as follows:

\vspace{3pt}\noindent\textbf{Definition 1.} \emph{Using an RGB-D camera, the robot must localize and grasp a target object described by a natural language expression.}\vspace{3pt}

However, ambiguity is a common issue in language grounding, with sources such as:

\vspace{3pt}\noindent\textbf{Assumption 1.} \emph{The target object is possibly unknown (i.e., novel objects) and partially observable (e.g., object occlusions, similar objects, etc.), and/or the query language is too vague to uniquely locate the target.}\vspace{3pt}

To resolve visual and linguistic ambiguities, we equip the grasping system with an attribute-guided POMDP planner for disambiguation.

\section{Methods}
\begin{figure*}[t]
  \centering
  \includegraphics[width=\textwidth]{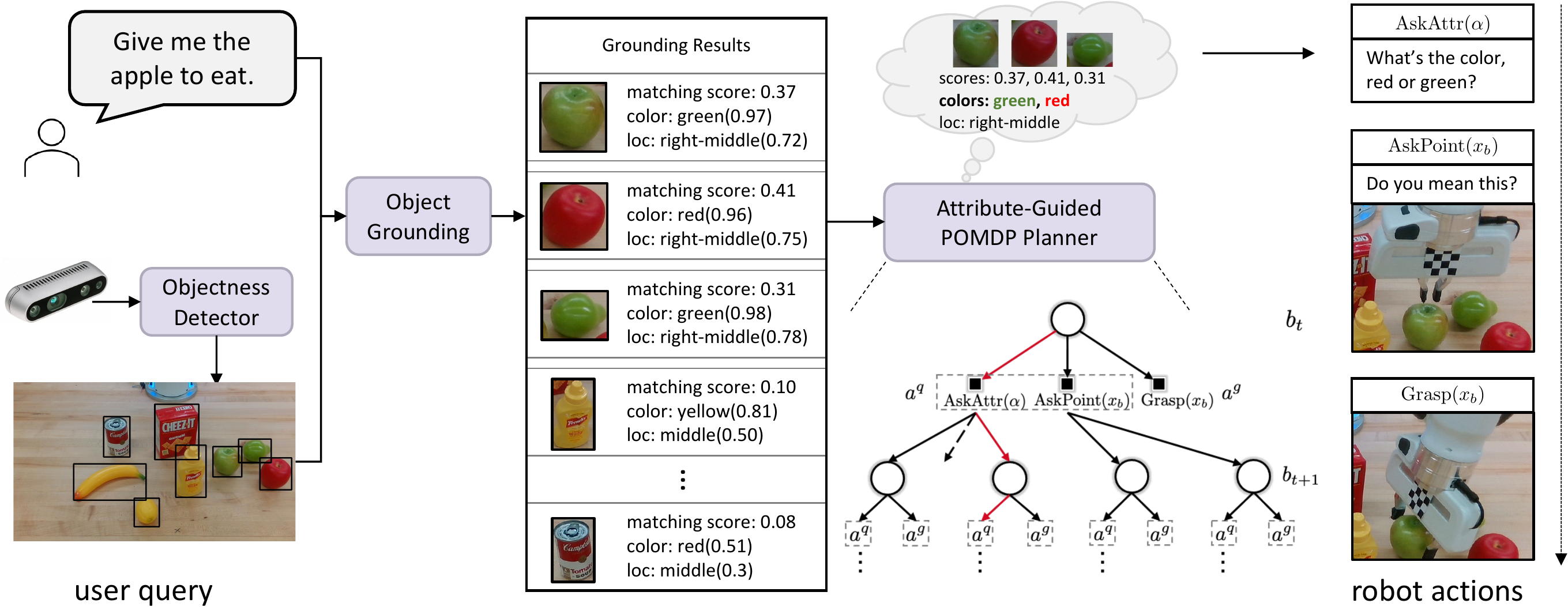}
  \caption{\textbf{Overview.} If a user wants the green apple but gives an ambiguous command, the red apple and green pear will cause target matching to fail. In our system, the object grounding module grounds each candidate object by predicting their matching score with the query language and the attributes. Using the grounding results as the observation model, the attribute-guided POMDP planner calculates the expected return of each asking action $a^q$ and grasping action $a^g$. The robot effectively resolves the ambiguity by asking attribute questions (about the color and location of the object) and pointing questions.}
  \label{fig:overview}
  \vspace{-10pt}
\end{figure*}
In our approach, the disambiguation module accepts the grounding results from an object grounding module and models the disambiguation process as an attribute-guided POMDP, as shown in Fig. \ref{fig:overview}. It is worth noting that the proposed disambiguation approach is independent of the grounding module and hence can be applied to improve the performance of any language grounding model.

\subsection{Object Grounding} \label{sec:method_a}
We adopt MAttNet~\cite{yu2018mattnet}, which uses CNN and LSTM encoders, as the grounding backbone. Given a query language $q$ and a set of candidate objects $\{x_i\}$, we train MAttNet on the RefCOCO dataset~\cite{yu2016modeling} to predict a matching score between the query and each candidate. To handle different query expressions (subject, location, and relationship), MAttNet has the modular design to compose query embeddings $[\mathbf{q}^{\text{subj}}, \mathbf{q}^{\text{loc}}, \mathbf{q}^{\text{rel}}]$ and object features $[\mathbf{x}_i^{\text{subj}}, \mathbf{x}_i^{\text{loc}}, \mathbf{x}_i^{\text{rel}}]$ for matching. The weighted average of the modular scores is used to calculate the overall matching score for a $(q, x)$ pair:
\begin{align}
    s(x | q) = \sum_{j \in \{\text{subj}, \text{loc}, \text{rel}\}} \omega^j \mathbf{q}^j \cdot \mathbf{x}^j
\end{align}

We ground the color and location attributes of each object in addition to the language. The color predictor $\phi_c$ (consisting of fully connected layers and the sigmoid function) takes as input the visual features $\{\mathbf{v}_i\}$ generated by the ResNet~\cite{he2016deep} encoder in MAttNet and predicts the probability of the 10 most common color values (e.g., red, green, blue, etc.). To prepare the color labels, we use the color words extracted from a template parser~\cite{kazemzadeh2014referitgame} which runs on the expressions of the RefCOCO dataset. For multi-color classification, a binary cross-entropy loss is used:
\begin{align}
    \mathbf{y}^c &= \phi_c(\mathbf{v})\\
    \mathcal{L}_{\text{c}} &= -\sum_{j} w_j \left[y_{j}^{\text{c}} \log p_{j}+(1-y_{j}^{\text{c}}) \log (1-p_{j})\right]
\end{align}
where $\mathbf{y}^c$ is the color probability from the predictor $\phi_c$, and $w_j = 1 / \sqrt{\text {freq}(\text{color}_j)}$ weights the color labels to compensate for unbalanced data~\cite{cui2019class}.

We summarize the object attributes for efficient disambiguation. Let $\mathcal{A}^\alpha \in \mathbb{R}^{n \times m}$ denote the attribute matrix of an attribute concept $\alpha \in \{\text{color, location} \}$ for $n$ objects and $m$ attribute states. With $\mathbf{e}_k$ as the $k$-th standard basis, we denote the attribute value vector for the $j$-th value (e.g., `red' for the color concept) as $\mathcal{A}^\alpha \, \mathbf{e}_j$ and the attribute state vector for the $i$-th object as $\mathbf{e}_i \mathcal{A}^\alpha$. The state vector $\mathbf{e}_i \mathcal{A}^\alpha$ represents a simplex probability distribution over the attribute states of the concept $\alpha$. As we consider the color and location attributes for disambiguation, the corresponding vectors are as follows:
\begin{align}
    \mathbf{e}_i \mathcal{A}^{\text{color}} &= \frac{\mathbf{y}^c}{\lVert\mathbf{y}^c\rVert_1}\\
    \mathbf{e}_i \mathcal{A}^{\text{loc}} &= \text{softmax}(-\lambda \mathbf{d}_i)
\end{align}
where $\mathbf{y}^c$ is the predicted color probability, $\mathbf{d}_i$ is the distance vector between the 9 location grids (e.g., top-left, top-middle, top-right, etc.) of the image and the center of the $i$-th object, and $\lambda$ is a numerical scale number.

\subsection{Attribute-Guided Disambiguation} \label{sec:method_b}
In our system, the disambiguation process is modeled as an attribute-guided POMDP (Attr-POMDP), wherein the agent acts to maximize its expected returns by ``imagining'' the outcomes of each action based on the grounding results.
\subsubsection{POMDP Model}
Our disambiguation POMDP~\cite{kaelbling1998planning} model is defined as a 6-tuple $<X, T, A, R, \Omega, O>$, where
\begin{itemize}
    \item The state space $X$ contains a set of candidate objects, in which the target object $x_d$ is hidden to the agent.
    \item The transition function $T(x, a, x')$ is deterministic because, we assume, the user's opinion of the target object does not change throughout the interaction.
    \item The action space $A$ consists of three types of actions: AskAttr($\alpha$), AskPoint($x_b$), and Grasp($x_b$). AskAttr$(\alpha)$ is an attribute-based question by which the agent asks to learn more about the attribute concept $\alpha$. We generate an attribute question by filling the decided concept and predicted values in a question template (e.g., ``what is the color of your target, red or yellow?''). AskPoint($x_b$) is a pointing-based question (e.g., ``do you mean this one?" accompanied by a physical pointing gesture) by which the agent inquires about the most likely object $x_b$ at the moment. The object $x_b$ is defined as the one with the maximum belief $b^t$ at time $t$, i.e., $x_b = \argmax_x b^t(x)$. Grasp($x_b$) is a physical action meaning the agent decides to grasp $x_b$ as the target object.
    \begin{center}
        \vspace{2pt}
        \begin{tabular}{c|c|c} 
        \hline
        $a$ & $x$ & $R(x, a)$ \\
        \hline
        AskAttr($\alpha$) & * & -0.1 \\
        AskPoint($x_b$) & * & -0.3\\
        Grasp($x_b$) & $x_b = x_d$ & 1 \\ 
        Grasp($x_b$) & $x_b \neq x_d$ & -1 \\ 
        \hline
        \end{tabular}
        \vspace{2pt}
    \end{center}
    \item The empirically designed reward function $R(x, a)$ motivates the agent to find the target $x_d$ as quickly as possible. As in the table above, we provide a large positive reward for grasping the target, a large negative reward for grasping an incorrect object, and smaller negative rewards for the questions AskAttr($\alpha$) and AskPoint($x_b$). The costs of the questions are set to approximately reflect the time of each action on a real robot.
    \item The observation space $\Omega$ contains all possible user responses to disambiguation questions.
    \item The observation function $O(x, a, o) = p(o | x, a)$ is the per-object conditional probability of the observation $o$ (a user’s verbal response) corresponding to a question.
\end{itemize}

\subsubsection{Observation Model}
While there are no restrictions on language expressions in our system, for efficiency, we sample question-related words\footnote{If a user chooses to give a complicated language response to our simple question, we simply re-run the language grounding module.} during planning, contrary to the vocabulary-level sampling in~\cite{whitney2017reducing} or the non-sampling in~\cite{shridhar2020ingress}. For an attribute question AskAttr$(\alpha)$, we sample words from the color or location vocabulary. The conditional observation probabilities for a sampled attribute value (e.g., `red') are obtained from the attribute matrix (see Sec. \ref{sec:method_a}):
\begin{align}
    p(o | x, a) = \mathcal{A}^a \, \mathbf{e}_o
\end{align}

For a pointing-based question AskPoint($x_b$), we sample binary responses. In our real-robot system, we determine if a user's response contains a word in the set of positive words $W_p=\{\text{`yes', ...}\}$ or the set of negative words $W_n=\{\text{`no', ...}\}$, as in~\cite{whitney2017reducing, shridhar2020ingress}. The conditional probability is
\begin{center}
    \begin{tabular}{c|c|c} 
    \hline
    & $o \in W_p$ & $o \in W_n$ \\
    \hline
    $x_i = x_d$ & 0.99 & 0.01\\
    $x_i \neq x_d$ & 0.01 & 0.99\\
    \hline
    \end{tabular}
\end{center}
where the values assume that a user is cooperative 99\% of the time and answers the questions truthfully.
\subsubsection{Belief Update}
To make action decisions sequentially, POMDP maintains a belief $b^t(x_i)$ over the candidate objects $\{x_i\}$. In our Attr-POMDP, the belief is initialized by the matching scores in Sec. \ref{sec:method_a} to reflect the initial confidence:
\begin{align}
    b^0(x_i) = \frac{\max(s(x_i | q), 0)}{\sum_k \max(s(x_k | q), 0)}
\end{align}

Every time the robot asks a disambiguation question and receives a response as the observation, it updates the belief using the observation model:
\begin{align}
    b^{t+1}(x)=\frac{1}{\mathbb{O}}\: p(o | x, a) b^t(x)
\end{align}
where $\mathbb{O}$ is the normalization term, and the objects with a higher probability of producing the observed response have a higher belief increase.

To solve the Attr-POMDP, we search a belief tree~\cite{somani2013despot}, as in Fig. \ref{fig:overview}, with each tree node representing a belief about the candidate objects. We search at each node to the depth limit $d=3$ by sampling actions, receiving observations, and updating the object belief. The goal of action planning is to achieve the best balance between the selection accuracy and the time cost, and the tree search yields an action with the highest expected return at the moment.

\subsection{Interactive Grasping System}
\begin{algorithm}[t]
\caption{Interactive Robotic Grasping System}
\textbf{Input}: RGB-D image $I$ and query language $q$
\begin{algorithmic}[1]
\State $\{x_i\} \gets \texttt{objectness\_detection}(I)$
\State $\{s_i\}, {\mathbf{y}^c_i} \gets \texttt{object\_grounding}(I, \{x_i\}, q)$
\State prepare $\mathcal{A}^{\alpha}$ using ${\mathbf{y}^c_i}$ and $\{x_i\}$
\State initialize belief $b^0(x_i)$ using $\{s_i\}$
\For{$t \gets 0$ to $\infty$}
\State $a = \texttt{pomdp\_plan}(b^t, \mathcal{A}^{\alpha})$
\If{$a$ == grasp}
\State Grasp($x_b$); break
\Else
\State $o = \texttt{ask}(\cdot)$ 
\State where $\texttt{ask}(\cdot) \in \{\text{AskAttr}(\alpha), \text{AskPoint}(x_b)\}$
\EndIf
\State $b^{t+1}\gets \texttt{belief\_update}(b^t, a, o)$
\EndFor
\end{algorithmic}
\label{alg:training}
\end{algorithm}
\begin{figure*}[t]
  \begin{subfigure}[t]{0.245\textwidth}
    \vskip 0pt
    \includegraphics[width=\textwidth]{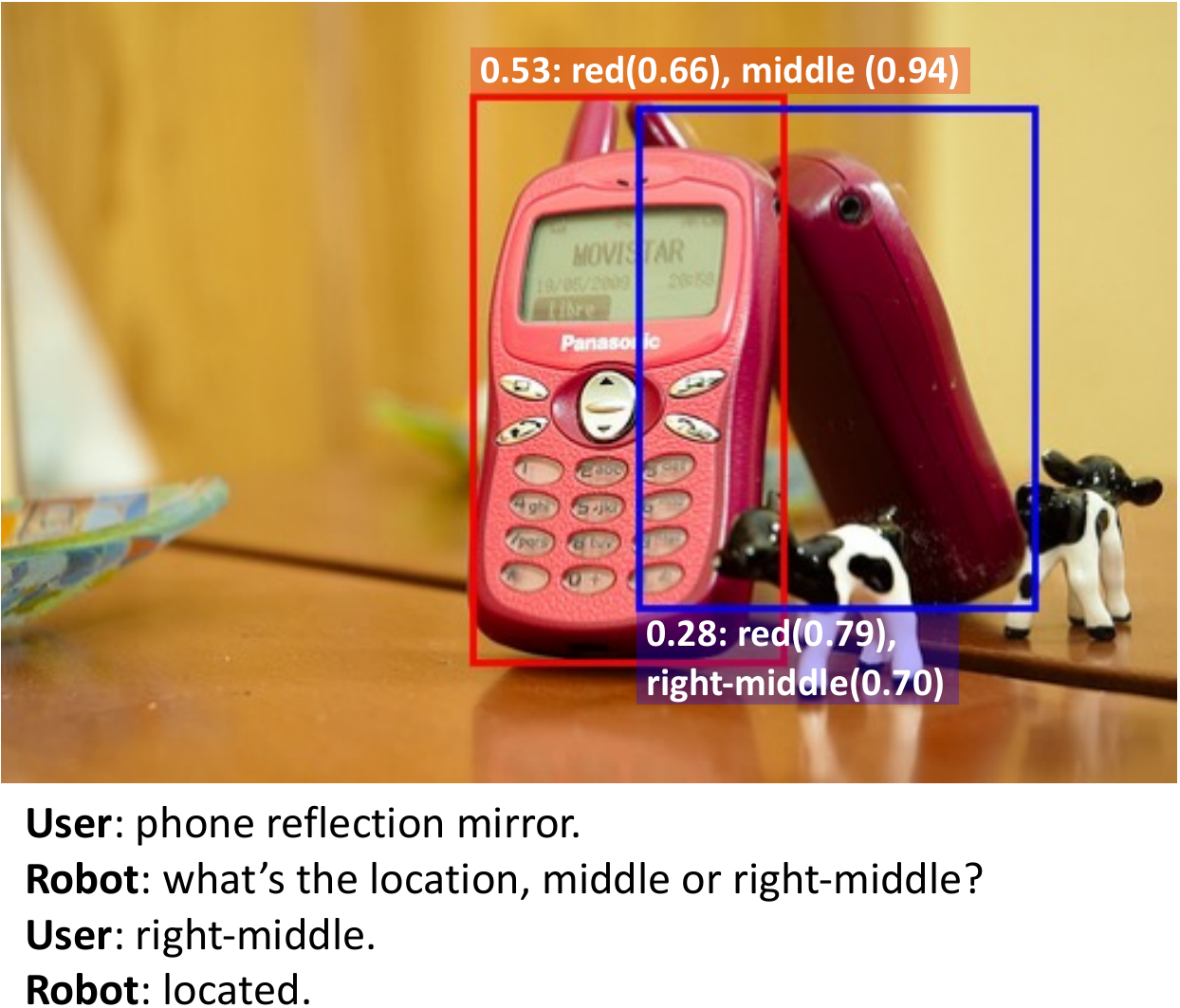}
  \end{subfigure}
  \begin{subfigure}[t]{0.245\textwidth}
    \vskip 0pt
    \includegraphics[width=\textwidth]{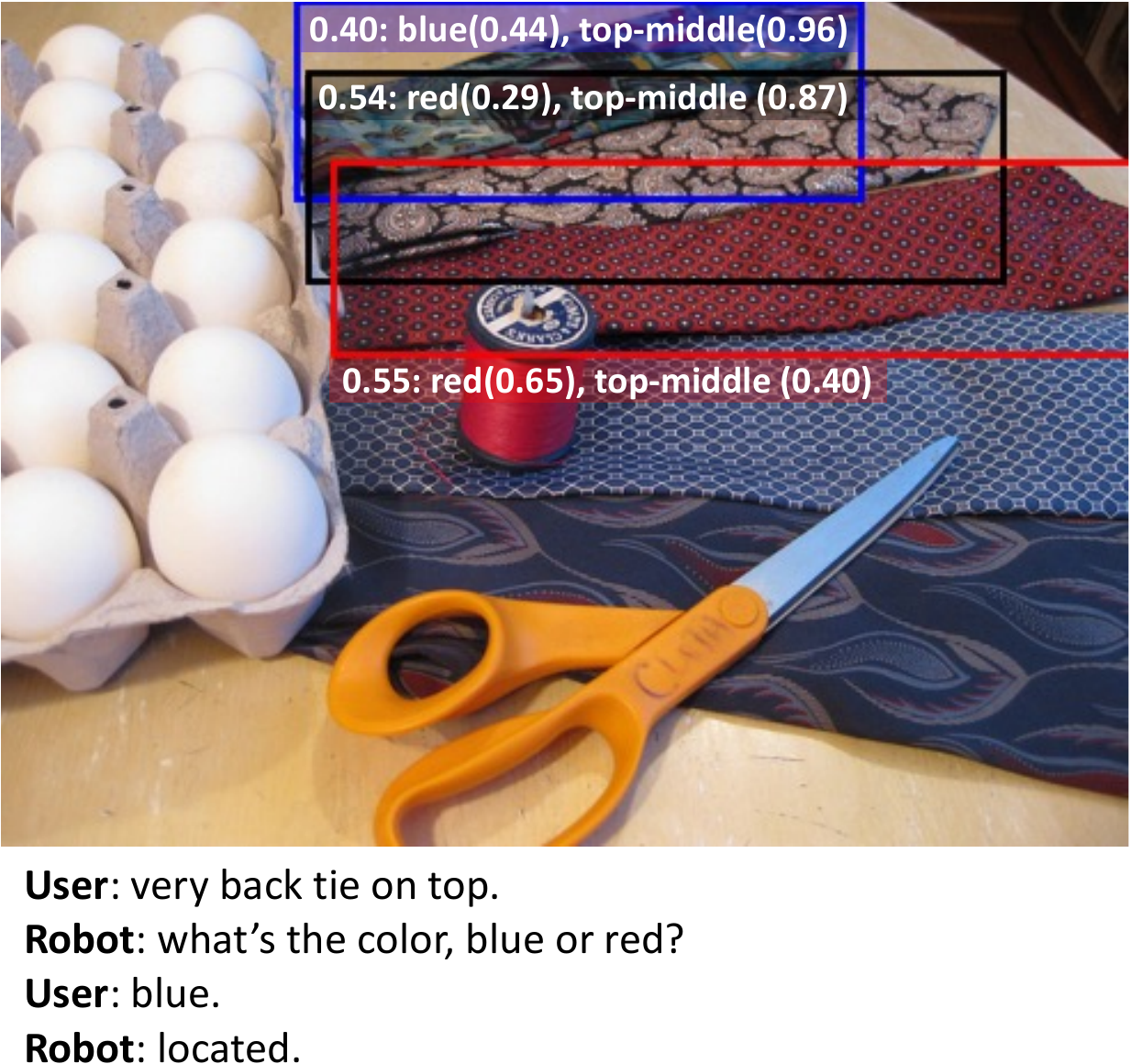}
  \end{subfigure}
  \begin{subfigure}[t]{0.245\textwidth}
    \vskip 0pt
    \includegraphics[width=\textwidth]{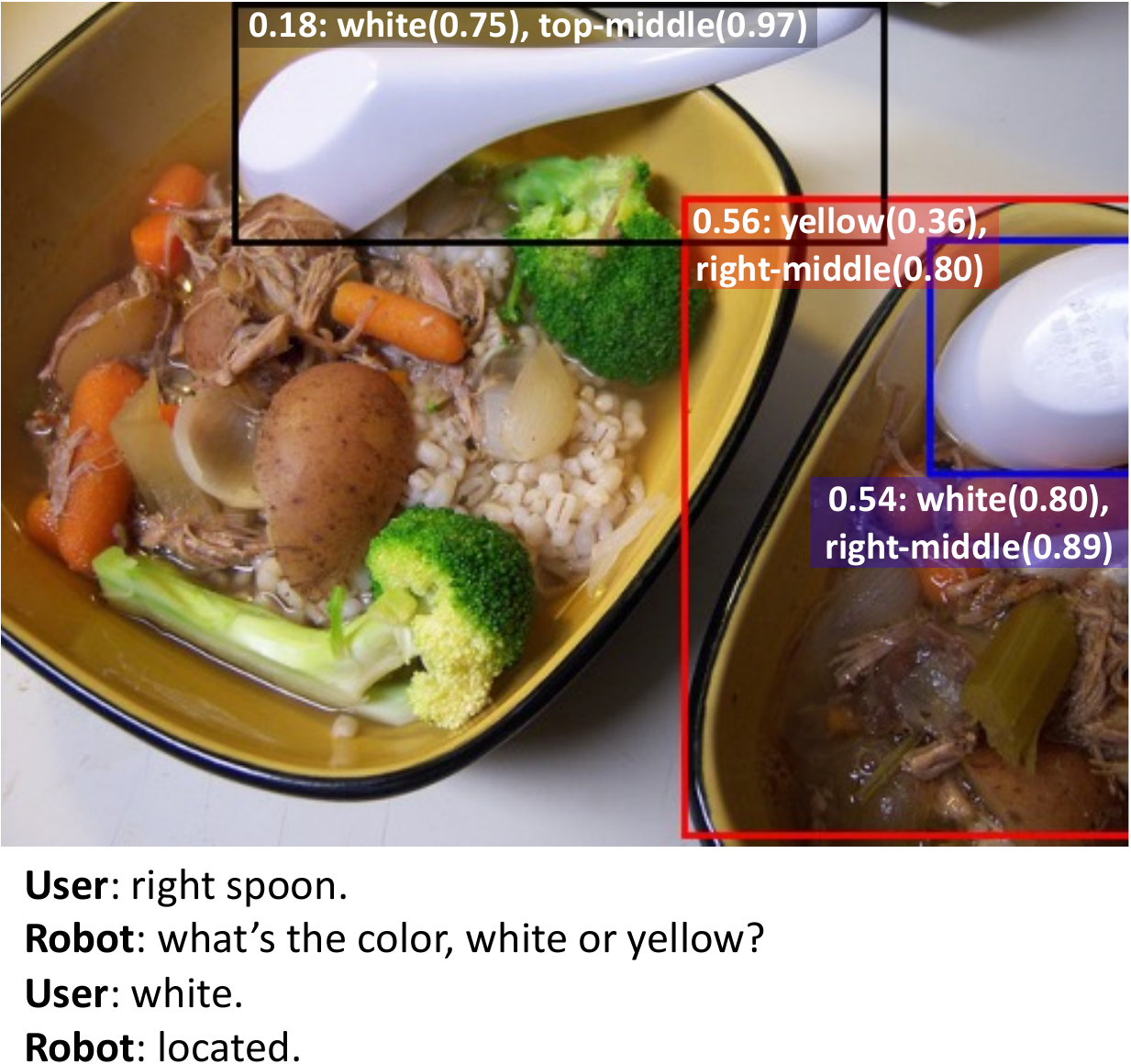}
  \end{subfigure}
  \begin{subfigure}[t]{0.245\textwidth}
    \vskip 0pt
    \includegraphics[width=\textwidth]{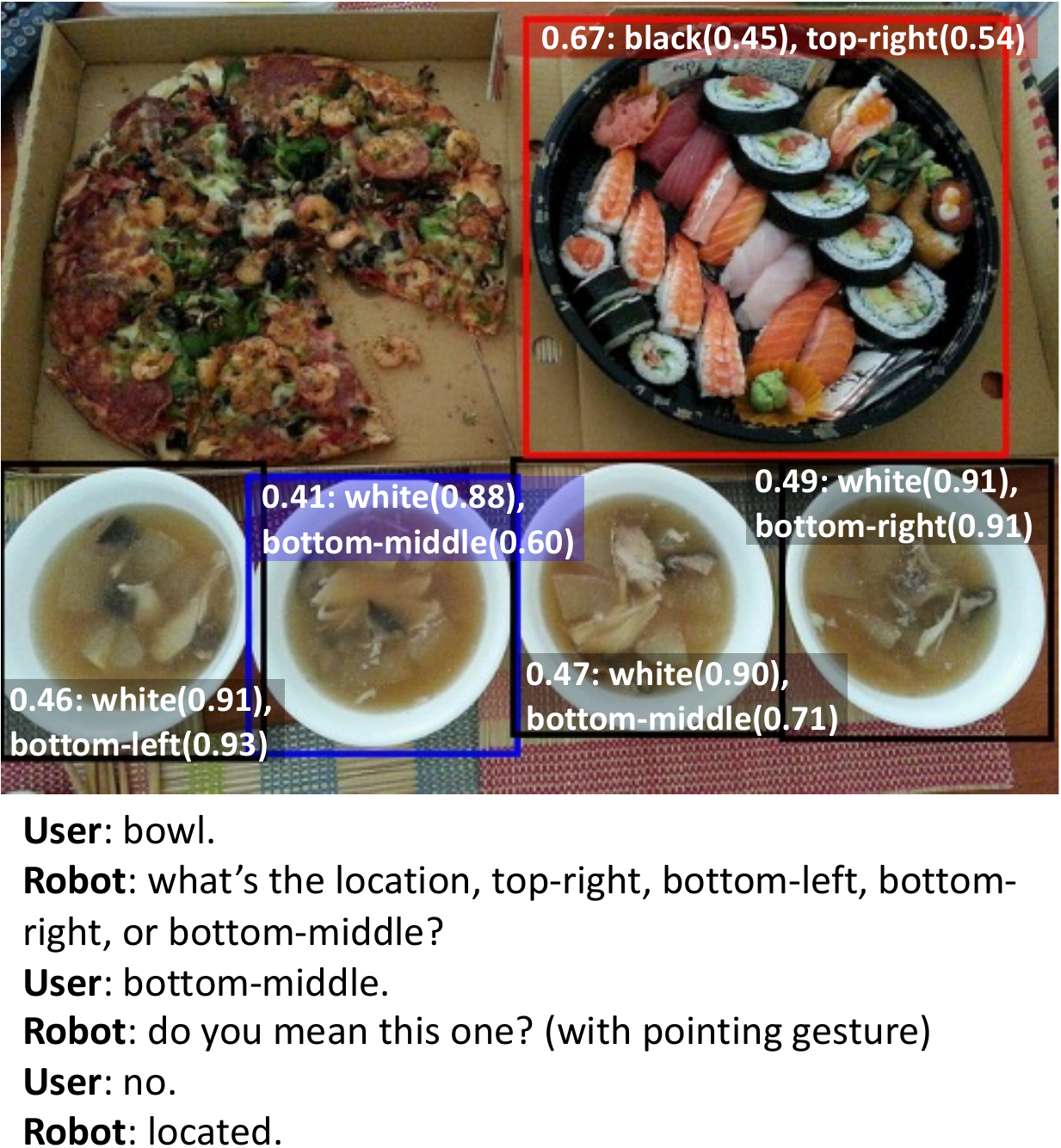}
  \end{subfigure}
  \vspace{-6pt}
  \caption{\textbf{Examples of disambiguation} on RefCOCO. In each case, we visualize the ambiguous objects with matching scores and outstanding attribute values. Blue bounding boxes denote referred target items, whereas red bounding boxes denote the most matched objects by the object grounding module. Our Attr-POMDP corrects the matching mistakes by asking attribute-based and pointing-based questions.}
  \label{fig:coco}
  \vspace{-12pt}
\end{figure*}
As shown in Fig. \ref{fig:overview} and Algorithm \ref{alg:training}, our interactive robotic grasping system comprises an objectness detector, the object grounding module in Sec. \ref{sec:method_a}, the disambiguation module in Sec. \ref{sec:method_b}, and a robotic manipulation module. We use the pre-trained UOIS-Net~\cite{xie2021unseen} to detect unknown objects in the real world. The predicted bounding boxes and RGB image are then fed into the object grounding module, with target matching scores and attribute matrices prepared for our Attr-POMDP disambiguation agent. Guided by object attributes, the agent sequentially plans attribute-based questions AskAttr($\alpha$), pointing-based questions AskPoint($x_b$), and object grasping Grasp($x_b$). When an object is selected for pointing or grasping, our manipulation module either plans a top-down pointing pose (based on the object location) or a 6-DOF grasp pose. The grasp poses are generated by the 3D CNN-based grasp generator in~\cite{lou2020learning},~\cite{lou2021collision}, which takes as input 3D point clouds of the selected object.

\section{EXPERIMENTS}
We executed a series of experiments to test the proposed approach against several baseline methods. The goals of the experiments are 1) to investigate how much the addition of the disambiguation module can improve the accuracy of the system, 2) to demonstrate that our Attr-POMDP can achieve a higher accuracy while asking fewer questions, and 3) to show the generalization of our system to novel objects and real-robot scenarios. We test the disambiguation performance of our Attr-POMDP on both RefCOCO images and a real robot. The interactive robotic grasping system is tested to retrieve and grasp a target object from a tabletop workspace. The entire system runs on a PC workstation with an Intel i7-8700 CPU and an NVIDIA 1080Ti GPU.

\subsection{Ablative Analysis} \label{sec:exp_a}
\begin{table}[b]
\vspace{-10pt}
    \centering
    \caption{Accuracy on RefCOCO (\%)}
    \vspace{-3pt}
    \label{tab:ablative}
    \begin{tabular}{c|c|c|c}
    \hline
    Query Type & RandSel & MAttNet & Ours\\
    \hline
    Unambiguous & 15.60 & 85.54 & \textbf{98.85} (w/\ 1.42 questions)\\
    Ambiguous & 15.60 & 27.34 & \textbf{92.69} (w/\ 1.71 questions)\\
    \hline
    \end{tabular}
    \vspace{8pt}
    \caption{Disambiguation Comparison on RefCOCO}
    \vspace{-3pt}
    \label{tab:disam}
    \begin{tabular}{c|c|c|c}
    \hline
    Method & Accuracy (\%) & \# Questions & Time(ms)\\
    \hline
    RandAsk & 80.28 & 3.46 & 0.12\\
    REG & 78.53 & 3.62 & 30.2\\
    FETCH-POMDP & 87.24 & 2.61 & 4023.8\\
    INGRESS-POMDP & 87.42 & 2.25 & 321.5\\
    Attr-POMDP(Ours) & \textbf{92.69} & \textbf{1.71} & 33.1\\
    \hline
    \end{tabular}
\end{table}
We perform an ablation study to validate our disambiguation module, comparing the accuracy with and without it. The methods are evaluated on the RefCOCO validation split ($\sim$ 1500 images), where the chance of randomly selecting an object (RandSel) is 15.60\%. With original unambiguous or ambiguous query expressions, an object in an image is referred to as the target. We report the results of the ablative analysis in Table \ref{tab:ablative}. When given unambiguous queries, the disambiguation module in our approach (MAttNet + Attr-POMDP) helps to correct the matching errors and significantly improves the accuracy from 85.54\% to 98.85\% with only 1.42 questions on average. For a thorough evaluation, we create challenging and ambiguous queries, one of which is either a category name (i.e., partial prior) or even none (i.e., no prior, and the belief is equally initialized). When given the ambiguous queries, the Attr-POMDP not surprisingly achieves an even larger accuracy improvement (from 27.34\% to 92.69\%) with a slightly larger number of questions. As illustrated in Fig. \ref{fig:coco}, our approach effectively resolves visual (e.g., similar objects and occlusion) and linguistic (i.e., vague query) ambiguities by asking questions. The Attr-POMDP tends to gather discriminative attribute information to rule out irrelevant objects, and points to the most likely object even when the designed attribute concepts are insufficient for disambiguation.

\subsection{Disambiguation Evaluation}
\begin{figure*}[t]
  \centering
  \includegraphics[width=\textwidth]{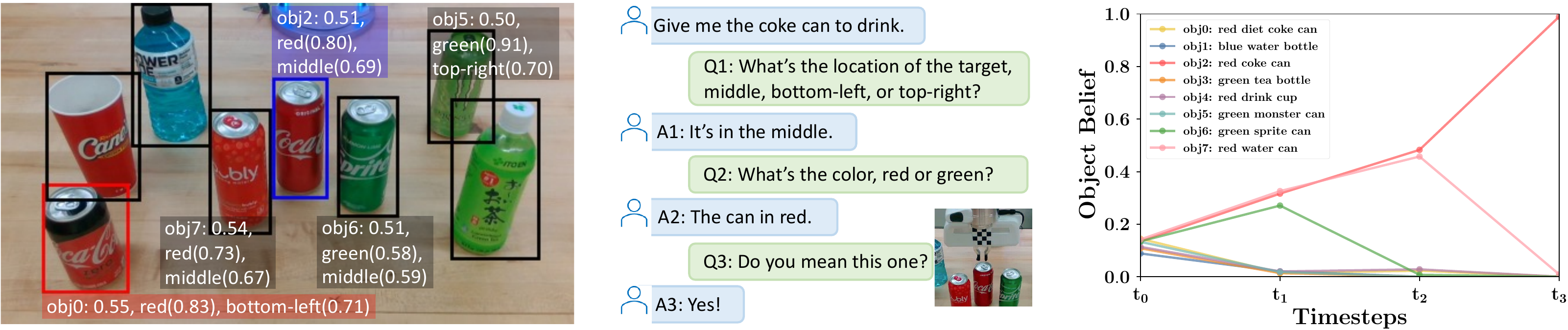}
  \caption{\textbf{Example of real-robot disambiguation.} A user wants the red original coke can (blue bounding box) despite other similar cans (e.g., red bounding box). By Attr-POMDP, the robot asks a sequence of questions (i.e., location, color, and pointing) and successfully retrieves the target. On the right, we visualize the object belief over timesteps, where the robot updates its belief upon receiving a response.}
  \label{fig:real}
  \vspace{-12pt}
\end{figure*}
We compare the disambiguation performance of Attr-POMDP with the following baselines: 1) \textbf{RandAsk} picks a random question from the action space of Attr-POMDP until it reaches the desired object belief or the maximum number of questions $n_q$. 2) \textbf{REG} asks about the most likely object using a language generator~\cite{mees2021composing} and receives a binary response indicating whether a generated question best matches the target. REG continues to ask until it receives a confirmation response or reaches $n_q$. 3) \textbf{FETCH-POMDP}~\cite{whitney2017reducing} and 4) \textbf{INGRESS-POMDP}~\cite{shridhar2020ingress} are other variants of POMDP (see Sec. \ref{sec:related} for more details). FETCH-POMDP exclusively uses pointing gestures to ask about candidate objects. INGRESS-POMDP extends FETCH-POMDP with spatial-semantic (e.g., red, middle, etc.) questions about candidate objects.

We evaluate the methods using the RefCOCO images and the ambiguous queries generated in Sec. \ref{sec:exp_a}. Based on the grounding results of MAttNet, each method is tested to find the target with the fewest number of actions. In Table \ref{tab:disam}, we report the accuracy, the number of questions asked, and the disambiguation time for each method. Overall, our approach significantly outperforms the baselines in both the accuracy and the number of questions. The accuracy improvement (from 27.34\% to 80.28\%) by RandAsk, which uses the actions of Attr-POMDP, validates the action space design of our approach. REG experiences limited generalization when captioning novel objects and performs slightly worse than RandAsk. Compared to the greedy-based approaches, the POMDP planning-based methods are capable of making action decisions sequentially and perform better in complex circumstances. Because of attribute guidance, our Attr-POMDP model achieves the highest accuracy (92.69\%) while asking the fewest number of questions (1.71) among the three POMDP models. Unlike FETCH-POMDP, which utilizes only pointing-based questions, and INGRESS-POMDP, which employs spatial-semantic questions, Attr-POMDP uses attribute questions to quickly rule out irrelevant objects before pointing. In addition, our Attr-POMDP largely outperforms the other two POMDP models in terms of disambiguation time. Having $O(n)$ actions at each tree node, both FETCH-POMDP and INGRESS-POMDP have $O(n^d)$ complexity, where $n$ is the number of objects and $d$ is the tree depth. On the contrary, the complexity of Attr-POMDP is only $O(c^d)$, where $c$ is constant, and thus its planning time does not scale with the number of objects.

\subsection{Real-Robot Experiments}
\begin{figure}[b]
\vspace{-8pt}
  \centering
  \includegraphics[width=0.48\textwidth]{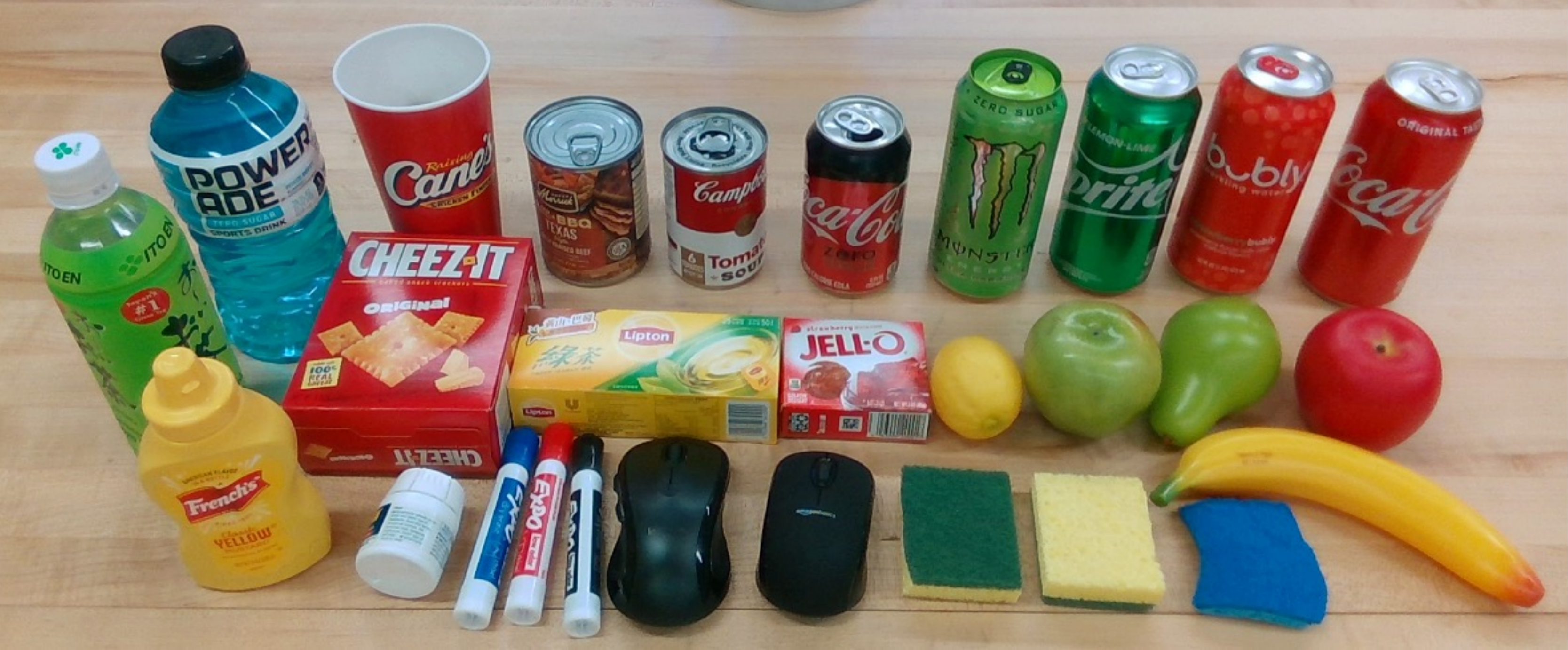}
  \caption{\textbf{Real-world testing objects.} We use various household objects to test our approach, some of which are confusing in terms of color, shape, or functionality.}
  \label{fig:real_obj}
  \vspace{-5pt}
\end{figure}
\begin{table}[!b]
    \centering
    \caption{Real-Robot Results}
    \label{tab:real}
    \begin{tabular}{c|c|c|c}
    \hline
    Method & Success\;/\;Trials & Accuracy (\%) & \# Questions\\
    \hline
    MAttNet & 32 / 70 & 45.71 & *\\
    RandAsk & 55 / 70 & 78.57 & 3.74\\
    REG & 52 / 70 & 74.29 & 3.84\\
    FETCH-POMDP & 57 / 70 & 81.43 & 3.20\\
    INGRESS-POMDP & 60 / 70 & 85.71 & 2.69\\
    Attr-POMDP(Ours) & \textbf{64} / 70 & \textbf{91.43} & \textbf{2.03}\\
    \hline
    \end{tabular}
\end{table}
The real-world configuration includes a Franka Emika Panda robot and an Intel RealSense D415 camera, statically mounted on a fixed tripod overlooking a tabletop scene. Fig. \ref{fig:real} shows an evaluation example of our interactive robotic grasping system in the real world. We use 28 real-robot testing objects, as shown in Fig. \ref{fig:real_obj}, some of which are never seen in the COCO dataset~\cite{lin2014microsoft}. For a fair comparison, we pre-generate 70 testing configurations consisting of object set, object pose, and query text for a target object. \revised{As a target query, an ambiguous language is generated by using object name, object relationship (e.g., ``on the left of''), and/or functionality (e.g., ``eat'') in the question template. The experiment participants are asked to respond to robot questions as consistently as possible, and we use a language parser to identify keywords in their responses.} The robot is required to retrieve and grasp the target within a combination of 8 $\sim$ 10 objects placed on the table. The trials, the accuracy, and the number of questions asked are all reported in Table \ref{tab:real}. The grounding module MAttNet only shows a 45.71\% accuracy with the input ambiguous queries, which necessitates using a disambiguation module. Overall, all disambiguation methods improve the accuracy in the real-robot experiments, though there are domain gaps. Using a language generator, REG's disambiguation performance is limited by the quality of language generation, particularly when there are novel objects and settings. Due to the influence of domain gaps on belief initilization and actions, the baseline planners FETCH-POMDP and INGRESS-POMDP show performance declines as well. In contrast, our approach, which makes use of generic object attributes and pointing gestures, shows better generalization. Among all the methods, our approach performs the best, achieving a 91.43\% accuracy. By planning its generic questions and grasping, Attr-POMDP efficiently increases the accuracy by over 45\% at the cost of asking only 2.03 questions on average, greatly outperforming the compared baselines. As shown in Fig. \ref{fig:real}, our approach updates the object belief sequentially by asking attribute-based and pointing-based questions.

\section{CONCLUSIONS}
In this work, we presented a novel interactive robotic grasping system with an attribute-guided POMDP (Attr-POMDP) for disambiguation. By using object grounding results as the observation model, the proposed Attr-POMDP model sequentially plans its asking and grasping actions, and effectively resolves visual and linguistic ambiguities by attribute-based and pointing-based questions. The Attr-POMDP planner significantly improves the accuracy of the backbone grounding module and is extensible to any other grounding model. We evaluated the system in both benchmark and real-robot settings. With ambiguous queries, our system achieved a 92.69\% accuracy on RefCOCO images and a 91.43\% accuracy in the real world. Our approach outperformed the other compared methods by large margins.

There are several avenues for future work. The current robotic system uses color and location attributes for disambiguation. It is highly anticipated to explore other attributes (e.g., size, shape, texture, etc.). Another direction is to acquire object attributes from dialogue with users, allowing for online knowledge update.

\bibliographystyle{IEEEtran}
\bibliography{references}

\end{document}